\documentclass[final,5p,times,twocolumn]{elsarticle}

\usepackage{amssymb}
\usepackage{amsmath}
\usepackage{algorithm}
\usepackage{algpseudocode}
\usepackage{booktabs}

\begin{document}

\begin{frontmatter}



\title{A Data-Efficient Sequential Learning Framework for Melt Pool Defect Classification in Laser Powder Bed Fusion}


\author[a]{Ahmed Shoyeb Raihan} 
\author[a]{Austin Harper}
\author[a]{Israt Zarin Era}
\author[a]{Omar Al-Shebeeb}
\author[b]{Thorsten Wuest}
\author[a]{Srinjoy Das}
\author[a]{Imtiaz Ahmed\corref{cor1}}
\ead{imtiaz.ahmed@mail.wvu.edu}

\address[a]{West Virginia University, Morgantown, WV 26505, USA}
\address[b]{University of South Carolina, Columbia, SC 29208, USA}

\begin{abstract}
Ensuring the quality and reliability of Metal Additive Manufacturing (MAM) components is crucial, especially in the Laser Powder Bed Fusion (L-PBF) process, where melt pool defects such as keyhole, balling, and lack of fusion can significantly compromise structural integrity. This study presents SL-RF+ (Sequentially Learned Random Forest with Enhanced Sampling), a novel Sequential Learning (SL) framework for melt pool defect classification designed to maximize data efficiency and model accuracy in data-scarce environments. SL-RF+ utilizes RF classifier combined with Least Confidence Sampling (LCS) and Sobol sequence-based synthetic sampling to iteratively select the most informative samples to learn from, thereby refining the model's decision boundaries with minimal labeled data. Results show that SL-RF+ outperformed traditional machine learning models across key performance metrics, including accuracy, precision, recall, and F1 score, demonstrating significant robustness in identifying melt pool defects with limited data. This framework efficiently captures complex defect patterns by focusing on high-uncertainty regions in the process parameter space, ultimately achieving superior classification performance without the need for extensive labeled datasets. While this study utilizes pre-existing experimental data, SL-RF+ shows strong potential for real-world applications in pure sequential learning settings, where data is acquired and labeled incrementally, mitigating the high costs and time constraints of sample acquisition.
\end{abstract}




\cortext[cor1]{Corresponding author. Tel.: +1-979-985-7439 ; fax: +1-304-293-4970.}

\end{frontmatter}



\section{Introduction}

Advanced manufacturing processes have emerged as a pivotal research domain, driven by innovative technologies that enhance production capabilities in line with the objectives of Industry 4.0. Among these technologies, Additive Manufacturing (AM) stands out for its transformative potential. AM offers unparalleled design flexibility, enables the rapid fabrication of complex geometries, minimizes material waste, and enhances performance reliability. Moreover, it supports the use of diverse materials, making it a versatile solution for modern manufacturing challenges \cite{ngo2018additive, ford2016additive}. While polymer-based AM printers are the most widely used, Metal Additive Manufacturing (MAM) has demonstrated significant potential in industries such as aerospace \cite{kumar2023metal}, automotive \cite{zhao2023direct}, and biomedical \cite{zhang2023application}, leveraging specialized alloys and materials. However, several challenges hinder the widespread adoption of MAM, including inconsistent printing quality and part defects \cite{wang2020machine, markl2016multiscale}. These issues compromise MAM’s ability to reliably deliver finished parts with the required mechanical properties and precise geometric accuracy. Addressing these limitations is critical to overcoming the barriers to industrial adoption and unlocking the full potential of this transformative technology.

Ensuring high-quality output in MAM depends on precise control and monitoring of the melt pool—the localized region where metal powder is melted and solidified by a directed energy source. Variations in process parameters, such as laser power, scan speed, hatch spacing, and layer thickness, influence melt pool behavior and, consequently, the quality of the final part. Defects such as lack of fusion, keyhole porosity, and balling often arise from irregularities in the melt pool, making it essential to monitor and adjust these parameters to reduce defect occurrence \cite{scime2019using}. In-situ monitoring techniques, which enable real-time data collection during the build process, play a pivotal role in capturing the intricate Process-Structure-Property (PSP) relationships that define MAM quality. By employing sensors like infrared cameras, photodiodes, and pyrometers, these techniques collect thermal and geometric data on the melt pool’s dynamic changes, offering valuable feedback to adjust process parameters on-the-fly. Such monitoring optimizes part quality and structural integrity without the need for costly post-process inspections \cite{ye2023predictions, herzog2024process}. Establishing a reliable classification model that links process parameters to melt pool defects provides a data-driven approach to understanding PSP relationships. This study aims to develop such a framework, facilitating the selection of optimal process settings that minimize defect formation and ultimately supporting MAM's goal of producing defect-free, high-performance parts \cite{gordon2020defect, wang2022data}.

There are two primary approaches to modeling PSP relationships in MAM: i) physics-based methods and ii) data-driven approaches \cite{kouraytem2021modeling}. Physics-based methods, which dominated early MAM research, rely on fundamental principles from thermodynamics, mechanics, and materials science to model the process. However, these approaches require extensive domain-specific knowledge and often involve complex multi-scale modeling, making them challenging to apply independently \cite{akbari2022meltpoolnet}. Data-driven approaches, which recognize patterns in datasets through techniques like regression, have increasingly gained prominence due to advances in Machine Learning (ML) and Deep Learning (DL) \cite{wang2022data} approaches. Innovations in computing, data collection, and algorithmic development have positioned ML and DL as cost-efficient alternatives to the high expense associated with experimental monitoring equipment required in physics-based methods \cite{akbari2022meltpoolnet}. Utilizing Artificial Intelligence (AI) enables improved MAM outcomes through enhanced part quality, cost-effective solutions, and optimized process parameters \cite{meng2020machine}. 

Despite their effectiveness, traditional ML and DL approaches face key limitations in melt pool defect classification, primarily due to the need for large experimental datasets, which are costly and time-intensive to obtain \cite{johnson2020invited}. These models depend heavily on the availability and quality of prior data, which is often limited, leading to imbalanced datasets where certain defect types are underrepresented, complicating training and often causing biased predictions \cite{du2020effects}. Additionally, melt pool characteristics vary significantly due to complex interactions among process parameters \cite{mukherjee2018mitigation}, requiring models to capture subtle differences in defect formation without overfitting to noise, which complicates feature extraction. Batch learning, commonly used in ML and DL, processes the entire dataset simultaneously during training, which can lead to inefficiencies by allocating equal importance to all samples, even though some may contribute less to improving the model's performance \cite{raihan2024augmented}. Furthermore, handling the high-dimensional data from thermal imaging and in-situ monitoring necessitates advanced feature extraction techniques, as standard ML and DL models struggle to generalize in data-scarce environments, limiting their capacity to capture the intricate PSP relationships crucial for accurate melt pool defect classification \cite{zhang2021prediction}.

To address the limitations of traditional approaches in MAM—such as the need for extensive labeled datasets, inefficiencies in batch learning, and challenges with high-dimensional, variable data—\textbf{Sequential Learning (SL)} presents a powerful alternative. SL is an iterative approach that strategically selects and queries only the most informative data points, significantly reducing the volume of labeled data required for training \cite{mojumder2023linking}. Unlike traditional learning, which requires all data to be labeled upfront, SL progresses in stages, enabling the model to refine its understanding by focusing on samples that maximize learning impact and minimize redundancy \cite{raihan2024augmented}. This approach is particularly effective in MAM, where data acquisition is costly, as SL prioritizes high-impact samples, thus reducing the need for exhaustive data collection. By adaptively selecting samples with the highest uncertainty or representativeness, SL can also address data imbalances, enhancing model robustness and generalizability. In melt pool defect classification, SL’s targeted sampling captures subtle distinctions in defect formation, crucial for managing the high variability in melt pool characteristics. Moreover, SL reduces computational inefficiencies inherent in batch learning by dynamically allocating resources to the most valuable data points, rather than treating all samples equally. SL’s adaptability to evolving data distributions makes it well-suited for handling high-dimensional data from thermal imaging and in-situ monitoring \cite{rovzanec2024active}. This sequential sampling strategy not only improves the model’s ability to establish PSP relationships with limited data but also enables progressive updates, enhancing classification accuracy and robustness without requiring extensive pre-collected datasets.

Keeping these considerations in mind, this study introduces the \textbf{SL-RF+} (Sequentially Learned Random Forest with Enhanced Sampling) framework—a novel SL model specifically crafted to optimize data efficiency in melt pool defect classification. SL-RF+ strategically addresses the limitations inherent in ML-based L-PBF defect classification through its integrated components as highlighted below:

\begin{itemize}
    \item \textbf{Sequential Learning Policy:} SL-RF+ adopts a SL approach that continuously adapts to new, informative samples as they are added, allowing the model to progressively improve its predictive accuracy. Sl-RF+ effectively mitigates the challenge of data scarcity, as the model can achieve robust performance with a fraction of the labeled data typically required by traditional methods.

    \item \textbf{Random Forest (RF) Classifier:} The RF classifier forms the backbone of SL-RF+ due to its effectiveness in handling high-dimensional, imbalanced datasets. Through ensemble learning, RF aggregates predictions from multiple decision trees, enhancing model stability and reducing sensitivity to variations in melt pool characteristics which is crucial for managing the variability in defect features observed in L-PBF.

    \item \textbf{Least Confidence Sampling (LCS):} Incorporating LCS allows SL-RF+ to prioritize uncertain samples, targeting those with the lowest confidence for labeling. This strategy improves learning efficiency, as selecting high-uncertainty samples accelerates model refinement and reduces the overall cost of data labeling—critical advantages in data-limited environments.

    \item \textbf{Sobol Sequence Sampling:} To thoroughly navigate the high-dimensional feature space, SL-RF+ employs Sobol sequences for quasi-random, uniform sampling across the parameter space. Sobol sequences provides diverse and evenly distributed sample points, enhancing the model's ability to generalize by capturing a broader range of defect patterns and reducing data gaps.
\end{itemize}

These components collectively make SL-RF+ a powerful and data-efficient framework tailored for the specific challenges of melt pool defect classification in MAM. The paper is organized as follows. Section 2 presents a comprehensive review of related work, covering both existing approaches for melt pool defect classification in MAM. Section 3 introduces the proposed SL-RF+ framework, detailing the integration of the RF classifier, LCS, and Sobol sequence sampling for efficient classification in data-scarce conditions. Section 4 describes the framework's application to the melt pool dataset. The results are presented and analyzed in Section 5, demonstrating the superior performance of SL-RF+ compared to conventional ML models. Finally, Section 6 concludes the paper by discussing the framework’s contributions, limitations, and potential real-world applications.

\section{Literature Review}

Classifying melt pool defects in Laser Powder Bed Fusion (L-PBF) is essential for ensuring MAM component quality and reliability \cite{chen2024meltpoolgan}. Various types of defects, including lack of fusion, balling, and keyhole formation, impact structural integrity and performance  of LPBF components \cite{snow2021toward}. \textit{Lack of fusion} occurs due to insufficient energy, causing poor layer adhesion and voids \cite{mukherjee2018mitigation}. \textit{Balling} results from surface tension issues, forming irregular droplets \cite{zoller2023numerical}. \textit{Keyhole} defects, from excessive laser power, lead to deep pores that weaken the structure \cite{yang2022quality}. Accurate defect classification is vital for optimizing process parameters and preventing failures, and various approaches have been explored to detect and classify these defects in real-time and post-process stages.

Traditional methods often rely on optical and thermal imaging techniques to capture real-time data on melt pool dynamics. Thermal imaging, for instance, can detect temperature fluctuations that correlate with defect formation, such as overheating in the case of keyholing or insufficient heating leading to lack of fusion \cite{clijsters2014situ}. X-ray computed tomography has been widely used as a post-process technique to identify and analyze internal defects, though it is costly and time-consuming \cite{du2020effects, liu2020study}. These imaging techniques, while effective, generate high-dimensional data, which can be challenging to interpret without advanced data processing and analysis methods \cite{zhang2021prediction}. Physics-based approaches, based on principles of heat transfer, fluid dynamics, and thermomechanics, aim to model melt pool formation in AM using analytical and numerical methods \cite{zhang2021prediction}. Analytical models, which derive closed-form solutions under simplified conditions, are computationally efficient but often fail to capture the complexities of the L-PBF process due to their assumptions \cite{angelastro2021integrated}. Numerical methods, such as Finite Element (FEM) and Finite Volume Methods (FVM), simulate melt pool behavior with higher accuracy by discretizing the geometry and solving partial differential equations iteratively \cite{soundararajan2021review}. Although these models handle complex geometries and material variations, they are computationally expensive and highly sensitive to mesh quality, boundary conditions, and material properties \cite{soundararajan2021review}. Despite their insights, the limitations in computational cost and accuracy of physics-based methods drive the need for alternative modeling approaches.

Data-driven approaches, which include ML and DL techniques, have gained popularity in AM due to their ability to model complex, nonlinear relationships without requiring explicit knowledge of the underlying physics \cite{guo2022machine}. These methods are particularly advantageous for modeling melt pool behavior in L-PBF, where a large number of process parameters influence melt pool characteristics. Traditional ML models, such as Support Vector Machines (SVM), Decision Trees (DT), K-Nearest Neighbors (k-NN), and ensemble methods like RF and Gradient Boosting (GB), have been widely applied to AM process modeling, including melt pool prediction \cite{akbari2022meltpoolnet}. ML models can effectively learn patterns from process parameters to predict melt pool properties like temperature, size, and shape, assuming large, comprehensive datasets are available. However, they are prone to overfitting when data is limited—a frequent challenge in AM, where labeled data is costly to obtain \cite{zhang2021prediction}. More recently, DL techniques, especially Convolutional Neural Networks (CNNs) and Recurrent Neural Networks (RNNs), have shown promise due to their ability to automatically learn hierarchical features from raw data \cite{ho2021dlam}. CNNs, which excel in image processing, have been applied to analyze thermal images of the melt pool, capturing spatial features that correspond to defect formations \cite{wang2024traditional}. RNNs are particularly effective for sequential data, enabling the modeling of temporal changes in melt pool characteristics by learning dependencies over time from data such as temperature profiles and melt pool dimensions \cite{abranovic2024melt}. The primary advantage of DL methods lies in capturing complex, nonlinear patterns without explicit feature engineering \cite{choudhary2022recent}. This makes DL models ideal for handling high-dimensional data from thermal imaging and time-series sensor data, where traditional ML models might struggle. Additionally, DL approaches are highly scalable and can achieve high accuracy with sufficient data, though they also require large labeled datasets and computational resources for effective training and inference, which can be a limitation in AM \cite{delgado2021deep}. Table~\ref{tab:ml_dl_meltpool_classification} showcases the broad use of data-driven techniques across various melt pool defect classification tasks.

\begin{table*}[!htb]
\centering
\caption{Summary of Existing Research on ML/DL in Melt Pool Defect Classification}
\label{tab:ml_dl_meltpool_classification}
\small
\begin{tabular}{ p{0.05\textwidth} p{0.60\textwidth} p{0.25\textwidth} }
\toprule
\textbf{Study} & \textbf{Description} & \textbf{Model Used} \\
\midrule
\cite{scime2019using} & Applied ML to detect keyhole and balling defects in L-PBF melt pools by analyzing morphology from in-situ images. & Unsupervised ML, Computer Vision \\
\cite{kwon2020deep} & Used DNN to classify melt pool images based on laser power to predict defect states. & Deep Neural Network (DNN) \\
\cite{dasari2020melt} & Analyzed melt pool dynamics and defect formation through process parameters and microstructural analysis. & Random Forests (RF)\\
\cite{gaikwad2022multi} & Combined temperature, shape, and spatter data for porosity detection, showing high accuracy with ML models. & SVM, RF, k-NN, CNN \\
\cite{ho2021dlam} & Developed DLAM for real-time porosity prediction in additive manufacturing with high accuracy. & Res-RCNN \\
\cite{mao2023continuous} & Proposed a method for continuous flaw detection using photodiode and temperature signals in LPBF. & BPNN, SSAE, LSTM \\
\cite{mahato2022detecting} & Used time-series data to monitor melt pool temperature and detect voids in PBF. & DTW, k-NN \\
\cite{gui2022detection} & Studied surface morphology and internal defects in PBF-EB, creating a process map for defect-free manufacturing. & SVM, Logistic Regression, Decision Tree, XGBoost, Naive Bayes \\
\cite{wang2023gaussian} & Developed a Gaussian Process model to classify melting states in L-PBF, enhancing process stability. & GPC, SVM, MLP, LSTM \\
\cite{yang2023defect} & Combined thermal and simulated images in a hybrid network for defect prediction, achieving high accuracy. & HNN, PSN \\
\cite{smoqi2022monitoring} & Used physics-informed melt pool features for porosity prediction with simple ML models, achieving high F1-score. & KNN, CNN \\
\cite{sato2024identification} & Applied k-means clustering to identify melt pool shapes and used XAI to interpret process parameters. & k-means, DNN, XAI \\
\cite{gu2024deep} & Leveraged EfficientNet b7 and DenseNet with U-Net, LinkNet, and FPN for accurate melt pool and porosity detection. & U-Net, LinkNet, FPN, EfficientNet b7, DenseNet 201 \\
\cite{khanzadeh2019situ} & Proposed in-situ monitoring with self-organizing maps for predicting melt pool porosity. & Self-Organizing Map (SOM) \\
\cite{ertay2021process} & Used high-dynamic range imaging and k-NN to classify melt pool states in DED. & k-NN \\
\cite{zhang2018extraction} & Employed melt pool, plume, and spatter features for anomaly classification in PBF. & PCA-SVM, CNN \\
\cite{wu2024situ} & Combined high-speed cameras, photodiodes, and microphones for real-time defect classification in LPBF. & Multi-Sensor Fusion, Improved LeNet-5 CNN \\
\cite{petrik2023meltpoolgan} & Introduced MeltPoolGAN to classify and generate melt pool images, achieving high accuracy. & Conditional GAN \\

\bottomrule
\end{tabular}
\end{table*}

While the techniques summarized in Table~\ref{tab:ml_dl_meltpool_classification} demonstrate promising results, a key challenge in L-PBF melt pool defect classification remains: the scarcity of large, labeled datasets. As discussed in the previous section, accurate defect labeling, especially for defects such as lack of fusion, balling, and keyhole formation, requires extensive and costly inspections such as X-ray CT or high-resolution optical microscopy \cite{du2020effects}. Limited data availability often leads to imbalanced datasets, complicating model training and potentially causing biased predictions that reduce model generalizability \cite{scime2019using}. Additionally, variability in melt pool characteristics—driven by interactions among process parameters like laser power, scan speed, and material properties-demands models be capable of distinguishing subtle variations without overfitting to noise \cite{mukherjee2018mitigation}. Traditional ML and DL approaches commonly employ batch learning, relying on one-shot training that requires all data to be labeled and available upfront. As noted earlier, this strategy can obscure valuable insights, since informative data points may be overshadowed by noisy or irrelevant samples, ultimately limiting the model’s generalization capability. Furthermore, the high-dimensional data generated through thermal imaging and in-situ monitoring in L-PBF necessitates sophisticated feature extraction techniques, which are challenging to implement effectively in data-scarce environments \cite{zhang2021prediction}. These limitations underscore the need for adaptive learning frameworks that can optimize data efficiency and improve model robustness \cite{ahmed2024toward}.

SL provides a solution by allowing models to selectively query the most informative samples, thus enhancing learning efficiency in data-scarce environments. By prioritizing uncertain or representative samples, SL reduces labeling costs and improves model robustness—an advantage particularly valuable in L-PBF applications where data acquisition is  expensive and labor-intensive. SL approaches are not new in the field of AM, and researchers have already employed various SL methods in their work. For instance, they have been applied in adaptive sampling for annotation reduction \cite{van2021active}, in-situ defect monitoring \cite{dasari2021active}, parameter optimization for porosity control \cite{mojumder2023linking}, visual quality control \cite{rovzanec2024active}, and experimental trials reduction in process optimization \cite{chepiga2023process, liu2022nonparametric}. However, despite the success of SL approaches in broader AM applications, there is a lack of specialized SL frameworks for melt pool defect classification in L-PBF processes. Our proposed SL-RF+ framework addresses this gap by introducing a tailored approach specifically designed to handle the unique challenges associated with L-PBF data—such as high dimensionality, limited availability, and class imbalance. This novel framework incorporates advanced sampling strategies and uncertainty quantification to pinpoint critical melt pool defects, achieving robust classification accuracy while minimizing the need for extensive labeled data. Our approach not only enhances data efficiency but also sets a new benchmark for defect classification in L-PBF by providing a performance standard against which traditional ML models reliant on large datasets can be evaluated.

\section{Methodology Overview}

In this section, before detailing our SL-RF+ framework, we discuss the foundational methods and concepts applied in this study, including Random Forest (RF) as the classification model, Least Confidence Sampling (LCS) as the sampling strategy, and Sobol sequence sampling for generating synthetic data points that enhance the exploration of the feature space.

\subsection{Random Forest Classifier}

The RF algorithm, introduced by Breiman~\cite{parmar2019review}, is an ensemble method that combines multiple decision trees to improve predictive accuracy and mitigate overfitting. This ensemble approach is particularly useful for high-dimensional and complex datasets. By aggregating predictions from multiple decision trees, each trained on a different subset of data, RF manages non-linear relationships effectively and provides robust performance, even in data-limited settings. RF operates based on ``bagging" principles, where each tree is trained on a bootstrapped sample of the original dataset, enhancing model diversity. For a test input \( x \), the final classification prediction \( \hat{y} \) is determined by majority voting across the trees in the forest:

\begin{equation}
    \hat{y} = \text{mode}(T_1(x), T_2(x), \dots, T_m(x))
\end{equation}

In this equation, \( T_j(x) \) represents the prediction from the \( j \)-th tree, and \( m \) is the total number of trees in the forest. The function \( \text{mode} \) takes the most frequent prediction among all trees, helping to smooth the decision boundary and reduce noise sensitivity. To increase tree diversity, RF randomly selects a subset of features at each split within each tree, reducing correlation among trees and creating a more stable model. For a feature set of \( p \) dimensions, typically \( k = \sqrt{p} \) features are randomly chosen at each split, which limits overfitting and makes the model robust to irrelevant or redundant features. The quality of a split is often measured by the Gini impurity metric \( G \), which assesses the ``purity" of a node:

\begin{equation}
    G = 1 - \sum_{c=1}^{C} p_c^2
\end{equation}

Here, \( C \) is the number of classes, and \( p_c \) denotes the proportion of samples of class \( c \) within the node. Lower Gini impurity indicates a purer node, meaning the samples within it are more likely to belong to a single class, thereby improving classification accuracy. The ability of RF to manage high-dimensional, non-linear data and its robustness against overfitting makes it well-suited for SL applications in L-PBF defect classification. By coupling RF with SL, our framework iteratively selects the most informative samples to label, maximizing data efficiency in scenarios where labeled data is limited and expensive to obtain.

\subsection{Least Confidence Sampling}

Least Confidence Sampling (LCS) is a SL strategy that iteratively selects the most uncertain samples for labeling, focusing on samples where the classifier is least confident. This approach is well-suited for defect classification in the L-PBF process, where labeled data are scarce~\cite{agrawal2021active}. In LCS, the confidence score for each unlabeled sample \( x \) is calculated as the maximum predicted probability for any class:

\begin{equation}
    \text{Confidence}(x) = \max_{y \in \{y_1, y_2, \dots, y_C\}} P(y|x),
\end{equation}

where \( P(y|x) \) is the probability of \( x \) belonging to class \( y \), and \( C \) is the number of classes. The sample \( x_{\text{LC}} \) with the lowest confidence score is chosen for labeling:

\begin{equation}
    x_{\text{LC}} = \arg \min_x \; \text{Confidence}(x).
\end{equation}

The Least Confidence Score (LCS) for a sample, representing its uncertainty, is defined as:

\begin{equation}
    \text{LCS}(x) = 1 - \max_{y \in \{y_1, y_2, \dots, y_C\}} P(y|x).
\end{equation}

Samples with the highest LCS values are prioritized for labeling, guiding the model to focus on regions of high uncertainty and effectively refining its decision boundary.

\subsection{Sobol Sequence Sampling}

To ensure thorough exploration of the feature space, Sobol sequence sampling is used in this SL framework. Sobol sequences, known for their low-discrepancy properties, provide quasi-random points that cover high-dimensional spaces uniformly~\cite{renardy2021sobol}. Each point \( s_i \) in a Sobol sequence is represented as:

\begin{equation}
    s_i = (x_{i1}, x_{i2}, \dots, x_{id}),
\end{equation}

where \( x_{ij} \) is the \( j \)-th dimension of sample \( s_i \). Sobol sequences distribute points systematically, minimizing gaps and clustering, making them suitable for assessing classifier uncertainty across the feature space. In our framework, Sobol sequences generate synthetic points that span the entire feature space, even in regions with limited labeled data. For each synthetic sample \( s_i \), the classifier predicts a probability vector, and the least confidence score \( LC(s_i) \) is computed as:

\begin{equation}
    LC(s_i) = 1 - \max(P(s_i)),
\end{equation}

where higher \( LC(s_i) \) values indicate greater uncertainty. The synthetic sample with the highest uncertainty serves as an ``ideal point" for exploration, and the closest real sample in the candidate pool is selected based on minimum Euclidean distance. This approach ensures efficient, informed exploration, minimizing the labeling burden by focusing on the most informative samples.

\subsection{Proposed SL-RF+ Framework}

In this section, the proposed SL framework (SL-RF+) is designed to optimize data efficiency by selectively labeling only the most informative samples. This framework integrates the strengths of RF classifier, LCS, and Sobol sequence-based synthetic sampling, forming a robust methodology for identifying and classifying defects with fewer labeled samples. SL-RF+ iteratively selects samples that the model is least confident about, allowing the classifier to focus on areas of uncertainty and refine its decision boundaries. By generating a broad set of synthetic samples using Sobol sequences, a comprehensive coverage of the feature space is ensured, thereby enabling the model to identify high-uncertainty regions that require further exploration. The SL-RF+ framework operates in the following steps:

\begin{enumerate}
    \item \textbf{Initial Data Partitioning and Model Training:} The dataset is initially divided into three subsets: $\mathcal{D}_{\text{initial}}$, used to train the RF classifier; $\mathcal{D}_{\text{candidate}}$, a pool from which samples are incrementally added in each iteration; and $\mathcal{D}_{\text{test}}$, a fixed test set for evaluating classifier performance. The RF classifier is first trained on $\mathcal{D}_{\text{initial}}$, establishing a baseline model on a small labeled dataset and setting the foundation for iterative improvement through SL.

    \item \textbf{Synthetic Sample Generation:} In each iteration, synthetic samples are generated across the feature space using Sobol sequence sampling. This quasi-random approach ensures comprehensive exploration of the feature space, aiding in identifying regions of high uncertainty.
    
    \item \textbf{Uncertainty Estimation via Least Confidence Sampling:} For each synthetic sample $s_i$ in the generated set, the classifier’s confidence score is computed as the maximum predicted probability for any class. LCS identifies the sample $s_{\text{ideal}}$ with the highest uncertainty  as the classifier's lowest confidence sample.
    
    
    \item \textbf{Selection of Informative Real Sample:} The ideal synthetic sample $s_{\text{ideal}}$ is matched to the closest real sample in $\mathcal{D}_{\text{candidate}}$ based on Euclidean distance. The nearest sample in feature space represents an informative, high-uncertainty instance from the real data.
    
    \item \textbf{Model Update and Iteration:} The selected real sample is added to $\mathcal{D}_{\text{initial}}$ for the next round of training, while it is removed from $\mathcal{D}_{\text{candidate}}$. The RF model is then retrained with the updated $\mathcal{D}_{\text{initial}}$, and the process repeats until the specified iteration budget is exhausted.
    
    \item \textbf{Performance Evaluation:} At each iteration, the model’s accuracy, precision, recall, and F1 score are computed on $\mathcal{D}_{\text{test}}$ to monitor the classifier’s improvement.
\end{enumerate}

This iterative framework effectively balances exploration and exploitation by strategically selecting samples from regions where the classifier is least certain, ensuring that the labeled dataset grows optimally with only the most informative samples. The framework incrementally builds a labeled dataset by iteratively selecting the most informative samples. The steps of the SL-RF+ framework is presented in Algorithm~\ref{alg:sl_rf_plus}.

\begin{algorithm}[!htb]
\caption{Proposed SL-RF+ Framework}
\label{alg:sl_rf_plus}
\begin{algorithmic}[1]
    \Require Initial dataset partition: $\mathcal{D}_{\text{initial}}$, $\mathcal{D}_{\text{candidate}}$, $\mathcal{D}_{\text{test}}$; iteration limit $N$; Random Forest (RF) classifier.
    \Ensure Trained RF model with optimized data efficiency.
    
    \State \textbf{Initial Training:} Train RF classifier on $\mathcal{D}_{\text{initial}} = \{\mathbf{X}_{\text{initial}}, \mathbf{y}_{\text{initial}}\}$.
    
    \For{$i = 1, 2, \dots, N$}
        \State \textbf{Generate Synthetic Samples:} Create a set of synthetic \State samples $\mathcal{S} = \{\mathbf{s}_j\}_{j=1}^{M}$ using Sobol sequence sampling 
        \State over feature bounds.
        
        \State \textbf{Calculate Least Confidence for Each Sample:}
        \For{each synthetic sample $\mathbf{s}_j \in \mathcal{S}$}
            \State Compute the classifier’s predicted probability vec-
            \State tor $\mathbf{P}(\mathbf{s}_j) = [P(y_1|\mathbf{s}_j), \dots, P(y_C|\mathbf{s}_j)]$.
            \State Calculate the confidence score: 
            \State $\text{Confidence}(\mathbf{s}_j) = \max_{y \in \{y_1, \dots, y_C\}} P(y|\mathbf{s}_j)$
            \State Compute Least Confidence Score (LCS) as: 
            \State $\text{LCS}(\mathbf{s}_j) = 1 - \max_{y \in \{y_1, \dots, y_C\}} P(y|\mathbf{s}_j)$
        \EndFor
        
        \State \textbf{Select Ideal Point:} Identify the synthetic sample with \State the highest uncertainty: $\mathbf{s}_{\text{ideal}} = \arg \max_{\mathbf{s}_j \in \mathcal{S}} \text{LCS}(\mathbf{s}_j)$
        
        \State \textbf{Select Real Sample from Candidate Pool:}
        \State Find the real sample in $\mathcal{D}_{\text{candidate}} = \{\mathbf{X}_{\text{candidate}}, \mathbf{y}_{\text{candidate}}\}$ 
        \State closest to $\mathbf{s}_{\text{ideal}}$: $\mathbf{x}_{\text{closest}} = \arg \min_{\mathbf{x} \in \mathbf{X}_{\text{candidate}}} \|\mathbf{x} - \mathbf{s}_{\text{ideal}}\|$
        \State Retrieve label $y_{\text{closest}}$ corresponding to $\mathbf{x}_{\text{closest}}$.
        
        \State \textbf{Update Training Set:} Add $(\mathbf{x}_{\text{closest}}, y_{\text{closest}})$ to $\mathcal{D}_{\text{initial}}$ 
        \State and remove it from $\mathcal{D}_{\text{candidate}}$.
        
        \State \textbf{Retrain Classifier:} Retrain RF model on the updated 
        \State $\mathcal{D}_{\text{initial}}$.
        
        \State \textbf{Evaluate Performance:} Test the classifier on $\mathcal{D}_{\text{test}}$ and 
        \State record performance metrics
    \EndFor
\end{algorithmic}
\end{algorithm}

\section{Application to Melt Pool Defect Classification}

\subsection{Melt Pool Classification Dataset}

\begin{figure}[!htb]
    \centering
    \includegraphics[width=\columnwidth]{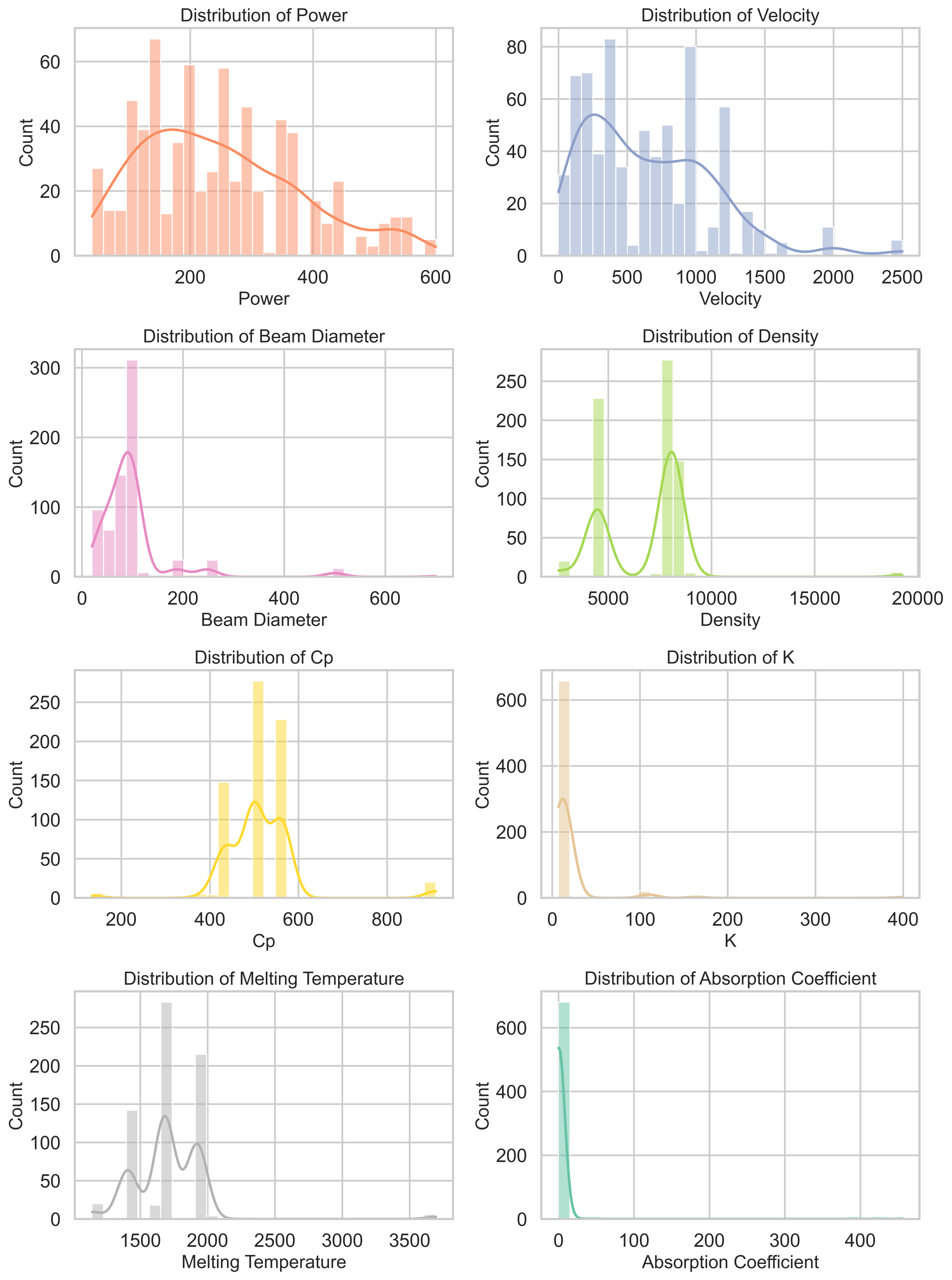}
    \caption{Distribution of input features in the melt pool dataset.}
    \label{fig:feature_distribution}
\end{figure}

The dataset used in this study pertains to melt pool defect classification in the L-PBF process, aiming to identify various defect types such as lack of fusion, balling, desirable, and keyhole defects \cite{akbari2022meltpoolnet}. It contains features relevant to the L-PBF process, which influence melt pool behavior and defect formation. These features include physical and process parameters like power, velocity, density, specific heat ($C_p$), thermal conductivity ($k$), melting temperature, beam diameter, and absorption coefficient. Table~\ref{tab:input_features_labels} provides a list of all the process parameters and the four classes of melt pool conditions along with their brief description. To understand the underlying characteristics of each feature, the distribution of all input features using histograms is analyzed in Figure~\ref{fig:feature_distribution}. These histograms highlight the diversity and range of values for each attribute, such as power, velocity, density, and others. The varying distribution shapes indicate that some features have a broader range, while others are more concentrated, which could affect the model's sensitivity to these attributes.

\begin{table}[!htb]
\centering
\caption{Input Features and Output Labels}
\label{tab:input_features_labels} 
\small 
\begin{tabular}{@{}p{0.37\columnwidth} p{0.58\columnwidth}@{}}
\toprule
\textbf{Feature / Label} & \textbf{Description} \\
\midrule
\multicolumn{2}{@{}l}{\textbf{Input Features}} \\
Power & Laser energy used for melting \\
Velocity & Speed of laser across the powder bed \\
Density & Material's mass per unit volume \\
Specific Heat ($C_p$) & Heat capacity of the material \\
Thermal Conductivity ($k$) & Heat conduction rate \\
Melting Temperature & Temperature where melting starts \\
Beam Diameter & Width of the laser beam \\
Absorption Coefficient & Fraction of laser energy absorbed \\
\midrule
\multicolumn{2}{@{}l}{\textbf{Output Labels}} \\
Desirable & Ideal melt pool with no defects \\
Lack of Fusion & Incomplete melting, weak bonding \\
Balling & Spherical defects causing uneven layers \\
Keyhole & Deep melt pool, potential for pores \\
\bottomrule
\end{tabular}
\end{table}

\begin{figure}[!htb]
    \centering
    \includegraphics[width=0.8\columnwidth]{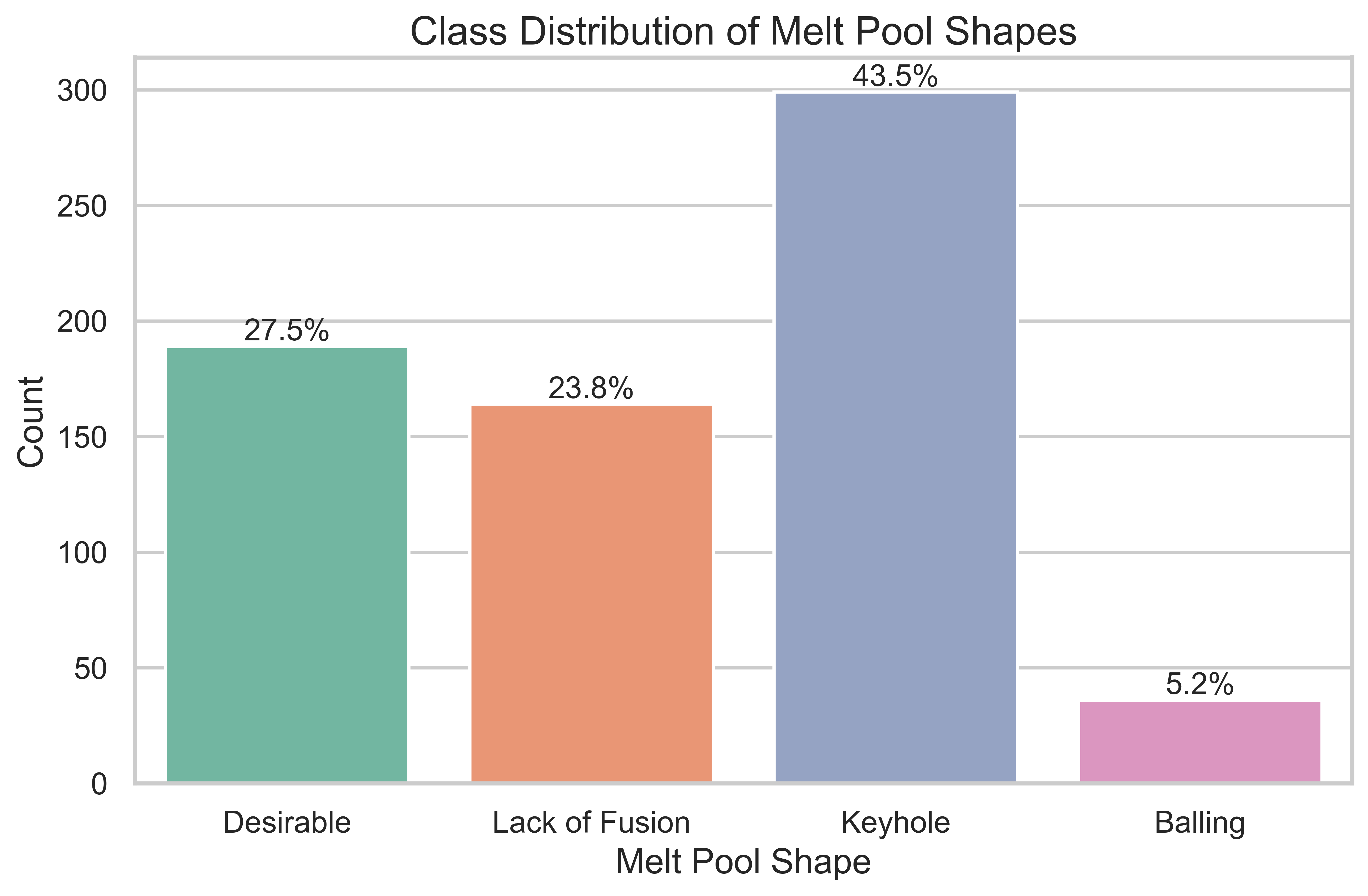}
    \caption{Class distribution of melt pool shapes in the dataset.}
    \label{fig:class_distribution}
\end{figure}

\begin{figure}[!htb]
    \centering
    \includegraphics[width=0.8\columnwidth]{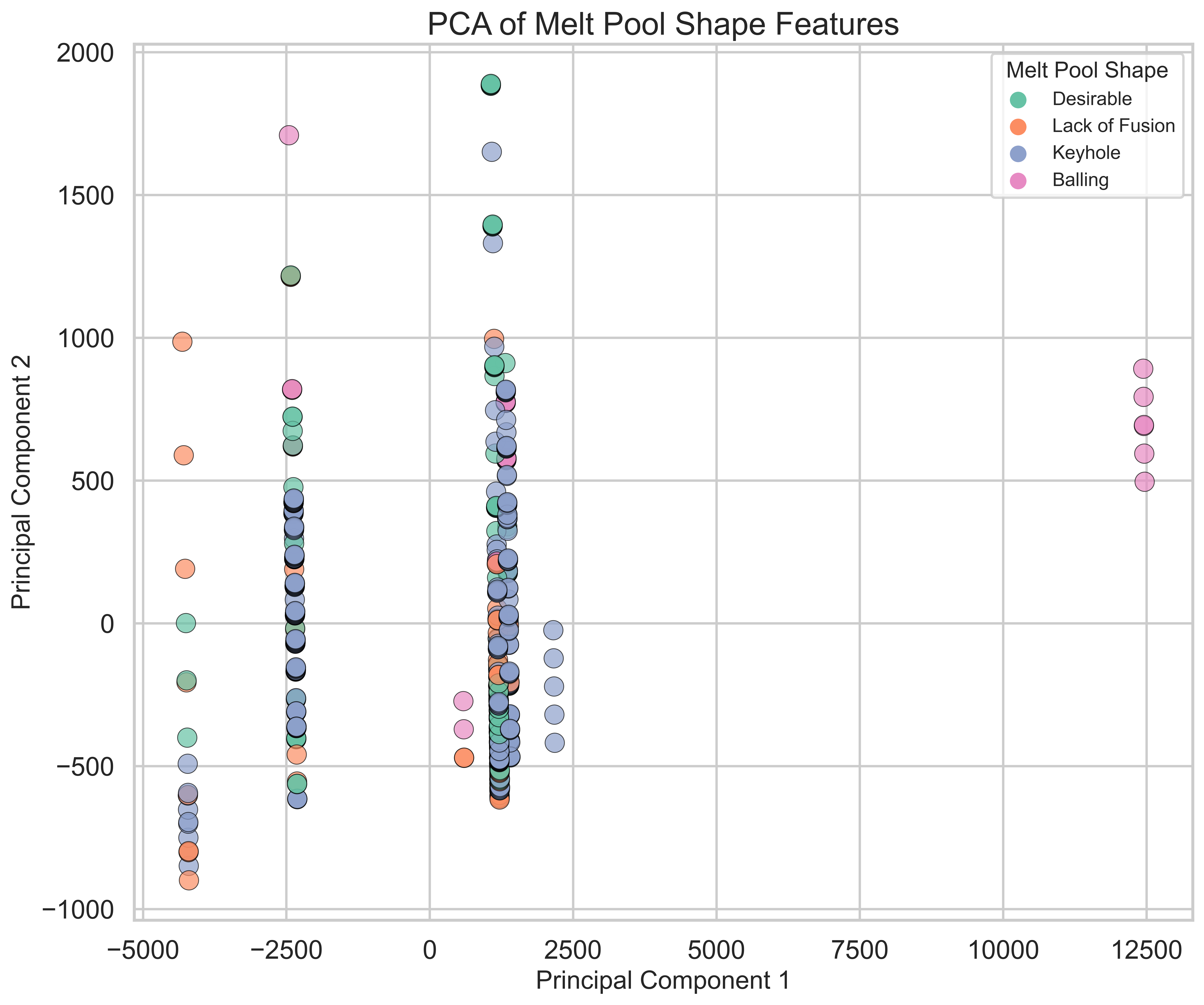}
    \caption{PCA plot showing the first two principal components of the melt pool shape features.}
    \label{fig:pca_plot}
\end{figure}

The class distribution of melt pool shape categories is presented in Figure~\ref{fig:class_distribution}. This plot shows that the ``Keyhole" class has the highest representation, followed by ``Desirable" and ``Lack of Fusion". ``Balling" has the least representation, which could pose class imbalance challenges during model training. Such imbalances often necessitate the use of specialized techniques to ensure fair representation and accurate classification across all classes. To assess the separability of classes based on feature space, Principal Component Analysis (PCA) is applied and the first two principal components were plotted as shown in Figure~\ref{fig:pca_plot}. The PCA plot provides insight into the data point's clustering behavior, revealing whether different classes form distinguishable groups in a reduced-dimensional space. While there is some overlap between classes, certain clusters are identifiable which indicates the features contain discriminative information that can assist the classifier in distinguishing defect types.

\subsection{Data Split and Sequential Sampling Process}

The dataset consisting of 685 samples has been divided into three main subsets for implementing the SL-RF+ framework: the initial training set $\mathcal{D}_{\text{initial}}$, the candidate pool $\mathcal{D}_{\text{candidate}}$, and the test set $\mathcal{D}_{\text{test}}$. Specifically, 25 samples were randomly selected for $\mathcal{D}_{\text{initial}}$ to train the initial Random Forest (RF) classifier. This small initial set allows the model to establish a preliminary understanding of the melt pool defect classification task. The remaining 460 samples are assigned to $\mathcal{D}_{\text{candidate}}$, which serves as the pool from which additional training samples are iteratively selected based on the model’s uncertainty. Finally, a separate test set, $\mathcal{D}_{\text{test}}$, consisting of 200 samples, has been reserved for evaluating the model's performance after each iteration.

To improve the robustness of results and reduce any uncertainty due to random fluctuations, the entire process was repeated for 30 independent runs. In each run, the dataset is split as described above and the SL-RF+ framework followed a sequential sampling strategy with a budget of 250 iterations, resulting in a total of 275 training samples by the end of the process ($\mathcal{D}_{\text{initial}}$ + 250 samples from $\mathcal{D}_{\text{candidate}}$). At each iteration, the model identifies and selects the sample with the highest uncertainty (measured via LCS) from a Sobol-generated synthetic sample set of 1000 synthetic samples. This ideal synthetic sample is then matched with the closest real sample in $\mathcal{D}_{\text{candidate}}$, which is then added to the training set. The RF model is retrained with the augmented dataset after each addition, gradually improving its understanding of melt pool defect patterns through a focused, data-efficient process. The hyperparameters of the RF classifier are optimized using a grid search with 5-fold cross-validation, ensuring robust performance and minimizing the risk of overfitting.

\section{Results}

To assess the effectiveness of the proposed SL-RF+ framework, several performance metrics are employed: accuracy, precision, recall, and F1 score. These metrics allow for a comprehensive evaluation of the model's ability to correctly classify various melt pool defect types across multiple performance dimensions. Additionally, to account for any performance variability due to random splits, each run is repeated 30 times, and the final metrics are averaged over these runs, to provide a reliable assessment of the model’s performance. For a comparative analysis, a baseline RF model where the same total number of 275 samples is used for training. However, in this traditional approach, all training samples are chosen randomly from the dataset without following a structured, sequential selection strategy. This approach allows for direct performance comparisons of the SL-RF+ framework with that of a standard RF classifier, highlighting the benefits of selecting informative samples iteratively. With this comparison method, we can quantify data efficiency and classification accuracy improvements achieved by leveraging SL.

\subsection{Performance Metrics Over Iterations}

\begin{figure*}[!htb]
    \centering
    \includegraphics[width=0.75\textwidth]{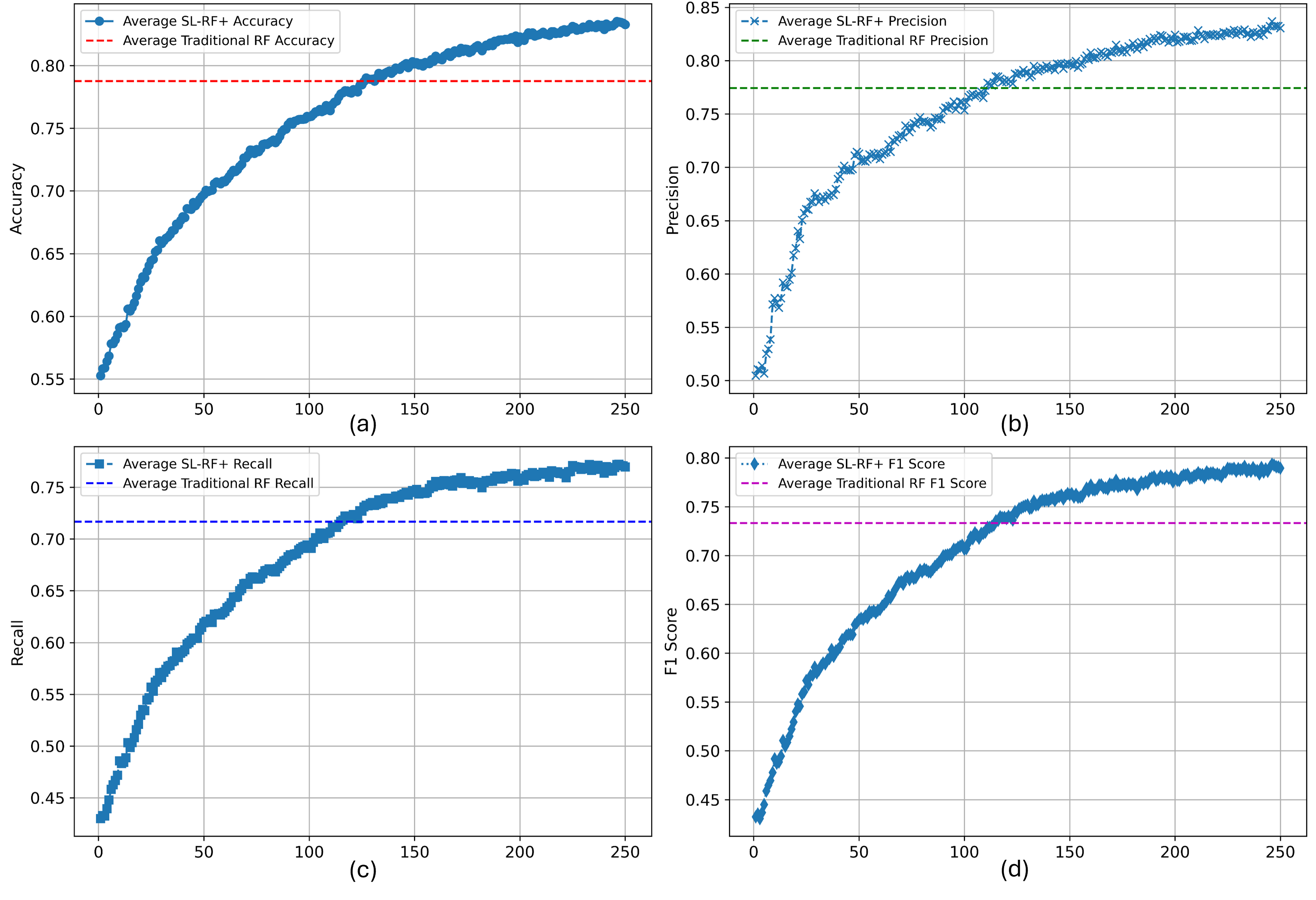}
    \caption{Evolution of performance metrics for SL-RF+ over 250 iterations for: (a) Accuracy, (b) Precision, (c) Recall, and (d) F1 Score}
    \label{fig:average_metrics}
\end{figure*}

The plots in Figure~\ref{fig:average_metrics} illustrate the progression of average performance metrics—accuracy, precision, recall, and F1 score—for the SL-RF+ model over 250 iterations. Each plot includes a dashed line indicating the performance benchmark of the traditional RF model trained on a full set of 275 samples. Notably, SL-RF+ achieves comparable performance to the traditional RF model between iterations 100 and 150 (seen from the intersection of the blue learning curve and the dashed lines), reaching similar accuracy, precision, recall, and F1 score metrics while using significantly fewer labeled samples. This efficiency stems from the SL strategy employed in SL-RF+, which combines LCS with Sobol sequence sampling to focus on the most informative samples. By selectively targeting high-uncertainty and diverse regions of the feature space, SL-RF+ accelerates its learning curve. This provides a significant advantage by allowing this framework to attain traditional RF-level performance well before reaching 275 labeled samples. In data-scarce environments, where collecting large datasets is expensive and time-consuming, the ability to achieve optimal performance with fewer samples translates to reduced costs and expedited model deployment. Additionally, the steady improvement in SL-RF+ metrics underscores its capacity to learn complex defect patterns efficiently, achieving comprehensive and robust classification without relying on a large labeled dataset.

\subsection{Confusion Matrix Analysis}

\begin{figure}[!htb]
    \centering
    \includegraphics[width=\columnwidth]{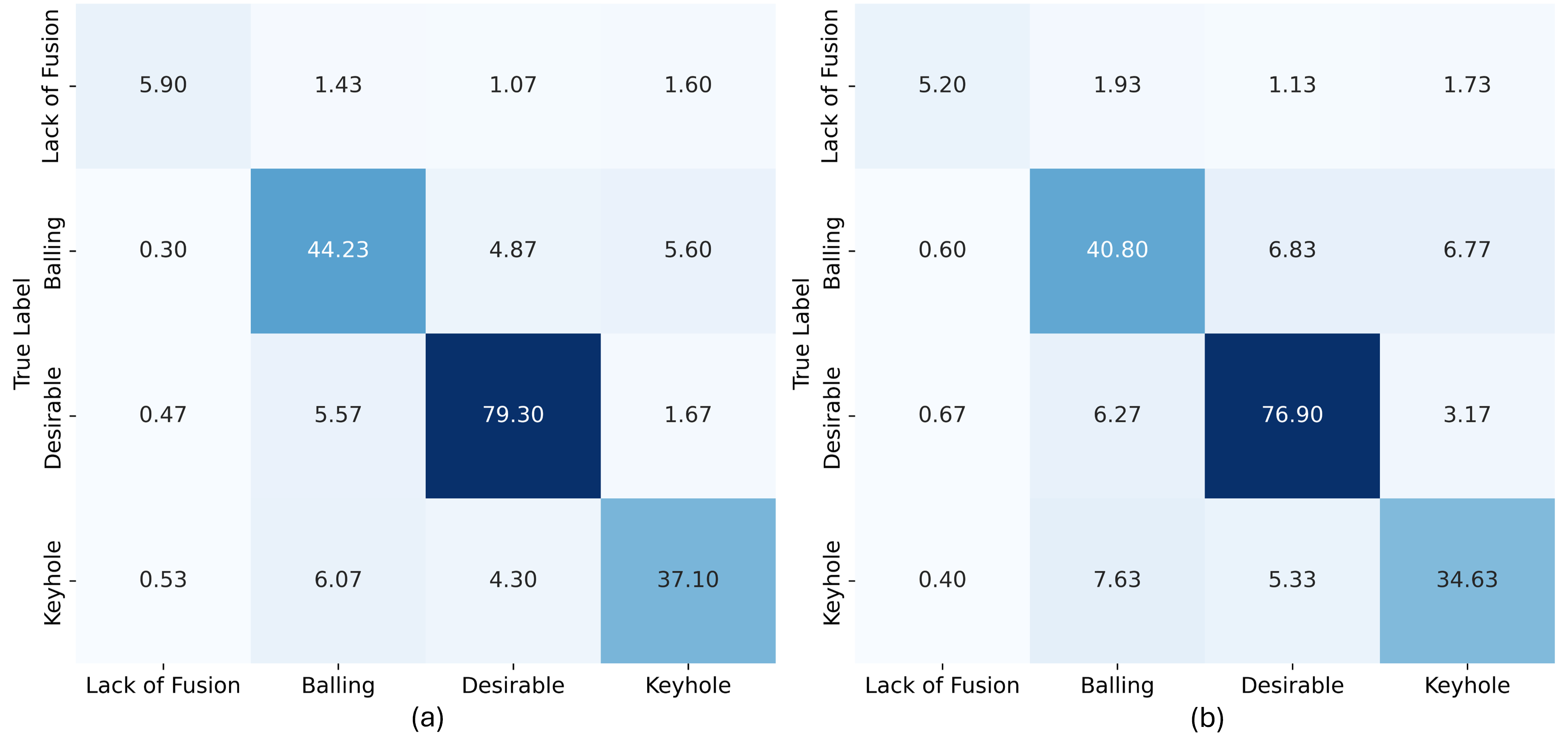}
    \caption{Confusion matrices for: (a) SL-RF+  and (b) Traditional RF}
    \label{fig:average_confusion_matrices}
\end{figure}

The confusion matrices averaged across 30 runs for the SL-RF+ model and the traditional RF model are illustrated in Figure~\ref{fig:average_confusion_matrices}. Each matrix provides an overview of the model's classification performance across four melt pool classes: ``Lack of Fusion", ``Balling", ``Desirable" and ``Keyhole". The values in each cell represent the average number of instances per class that were correctly or incorrectly classified, averaged across 30 runs. In the SL-RF+ matrix (Figure~\ref{fig:average_confusion_matrices}a), higher values are observed along the diagonal, particularly for the ``Desirable" and ``Balling" classes, indicating strong classification accuracy for these categories. When compared to the baseline RF model (Figure~\ref{fig:average_confusion_matrices}b), the sequential RF shows marginally higher values in the true class predictions across most classes. Notably, SL-RF+ performs slightly better in distinguishing ``Balling" from other classes, which is typically challenging due to class imbalance. This improved accuracy highlights the benefit of the SL approach in focusing on uncertain samples, which refines the model’s ability to differentiate between classes with limited labeled data.

\subsection{Class-wise Performance Comparison}

\begin{figure*}[!htb]
    \centering
    \includegraphics[width=0.7\textwidth]{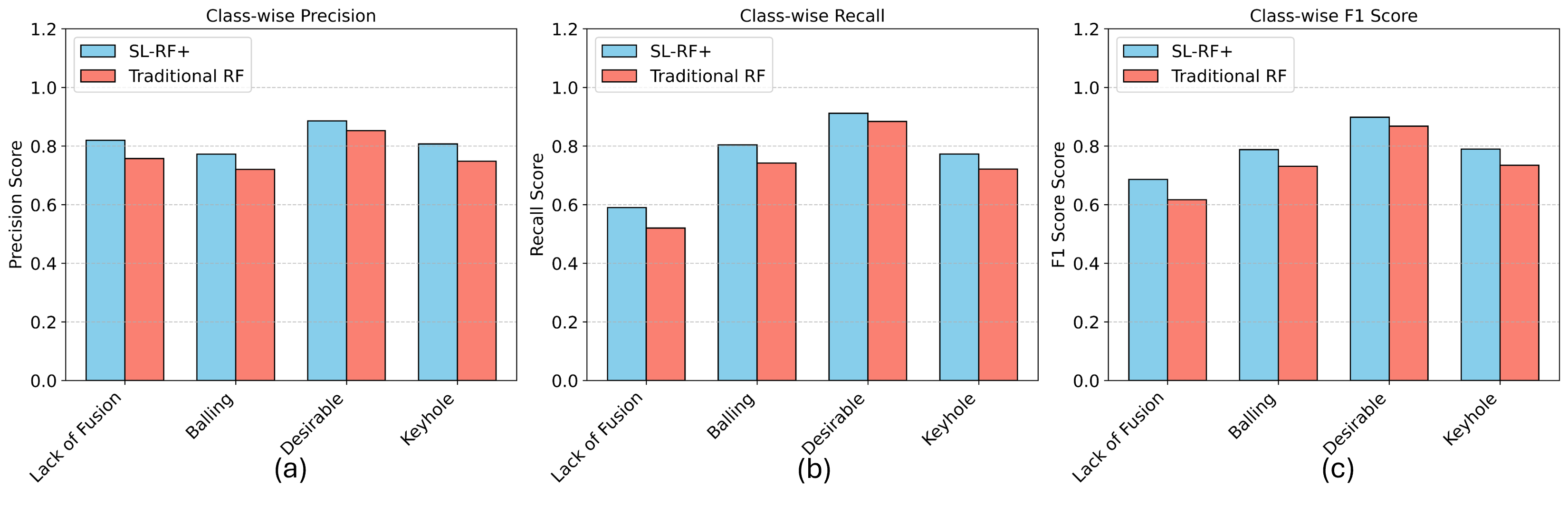}
    \caption{Class-wise performance comparison between SL-RF+ and traditional RF models for: (a) Precision, (b) Recall, and (c) F1 Score}
    \label{fig:class_wise_performance}
\end{figure*}

Figure~\ref{fig:class_wise_performance} provides a comparative analysis of the class-wise precision, recall, and F1 Score for the SL-RF+ and traditional RF models. SL-RF+ exhibits slightly higher precision scores (Figure~\ref{fig:class_wise_performance}a) across all classes compared to the standard RF, with notable improvements for the ``Balling" and ``Keyhole" classes, indicating a reduction in false positives for these challenging defect types. In terms of recall, as seen in Figure~\ref{fig:class_wise_performance}b, the sequentially learned RF model shows enhancements, particularly for the ``Lack of Fusion" and ``Balling" classes, suggesting improved identification of true instances of these defects, which are harder to classify due to limited representation. The F1 Score, which balances precision and recall, is higher for the SL-RF+ model across all classes, with significant improvements observed for the ``Balling" and ``Keyhole" classes (Figure~\ref{fig:class_wise_performance}c). These class-wise metrics collectively illustrate that the SL-RF+ framework not only enhances overall classification performance but also demonstrates substantial gains in identifying and classifying less frequent defects. These improvements are critical for effective quality control in melt pool defect classification.

\subsection{Effect of Training Sample Size}

\begin{figure}[!htb]
    \centering
    \includegraphics[width=\linewidth]{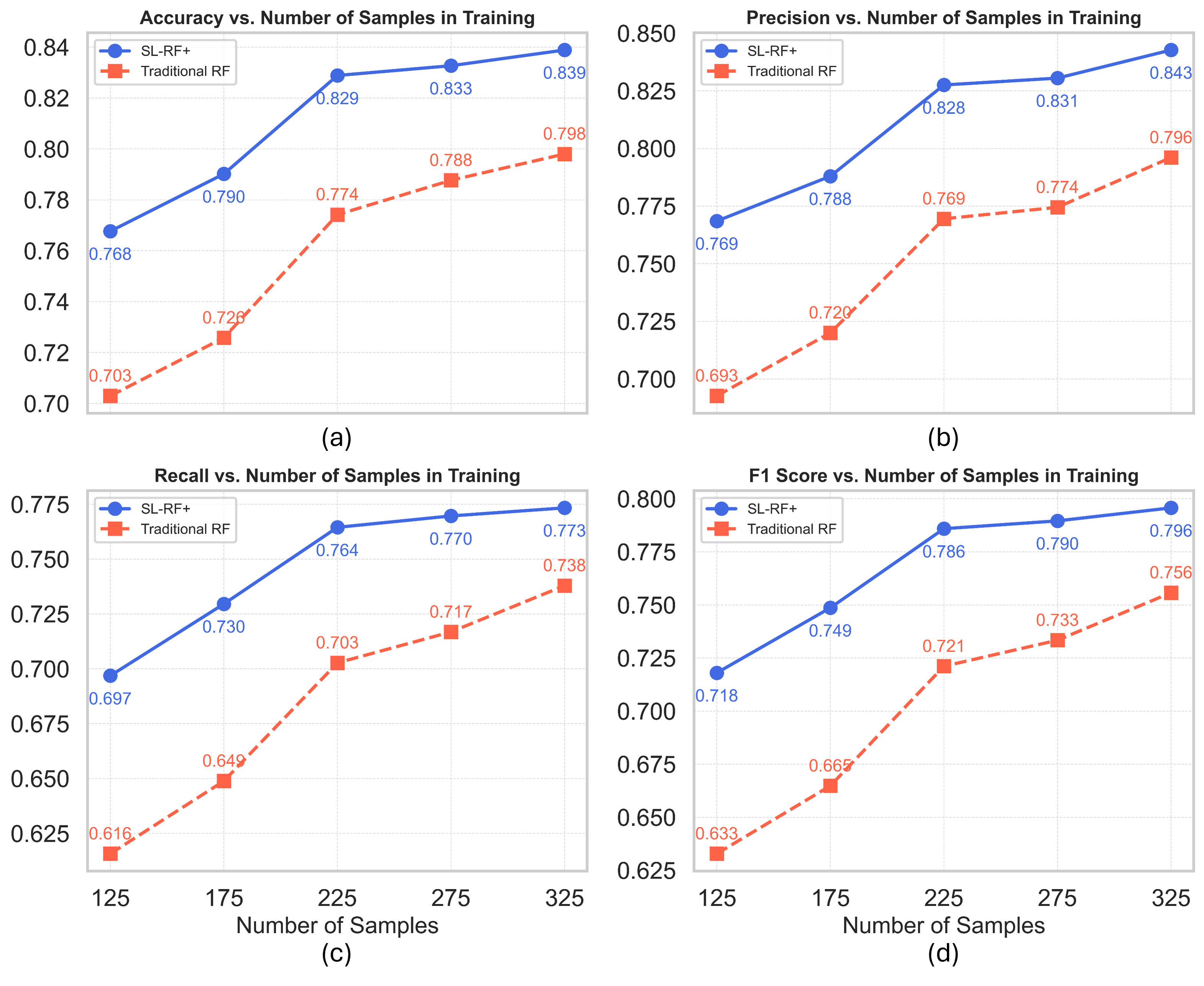}
    \caption{Effect of training sample size on performance metrics for SL-RF+ and traditional RF for: (a) Accuracy, (b) Precision, (c) Recall, and (d) F1 Score}
    \label{fig:changing_training_samples}
\end{figure}

The plots in Figure~\ref{fig:changing_training_samples} illustrate the effect of training sample size on the performance metrics—accuracy, precision, recall, and F1 score—of the SL-RF+ model compared to the traditional RF. These plots clearly show that SL-RF+ achieves superior or comparable performance with fewer samples due to its SL strategy. For instance, in Figure~\ref{fig:changing_training_samples}a, the accuracy of SL-RF+ with only 175 training samples (25 initial + 150 sequential) reaches 0.790, surpassing the accuracy of traditional RF, which achieves 0.788 but requires 275 training samples. A similar trend is observed for other metrics: in Figure~\ref{fig:changing_training_samples}c, SL-RF+ attains a recall score of 0.730 with 175 samples, outperforming traditional RF’s recall of 0.717 achieved with 275 samples. Notably, even with 175 samples, SL-RF+ demonstrates competitive recall performance (0.730) close to traditional RF’s recall (0.738) obtained with 325 samples, an additional 150 samples. These results emphasize the value of SL in efficiently selecting the most informative samples, allowing SL-RF+ to achieve comparable or better performance with significantly fewer samples. Another notable observation is the initial rapid improvement in SL-RF+ metrics, which gradually slows as more samples are added. This trend is due to the model's ability to initially prioritize the most informative samples from the candidate pool \( \mathcal{D}_{\text{candidate}} \), resulting in quicker learning gains that taper as more samples are incorporated.

\subsection{Sequential Learning Strategy in Other ML Models}

\begin{figure}[!htb]
\centering
\includegraphics[width=0.9\columnwidth]{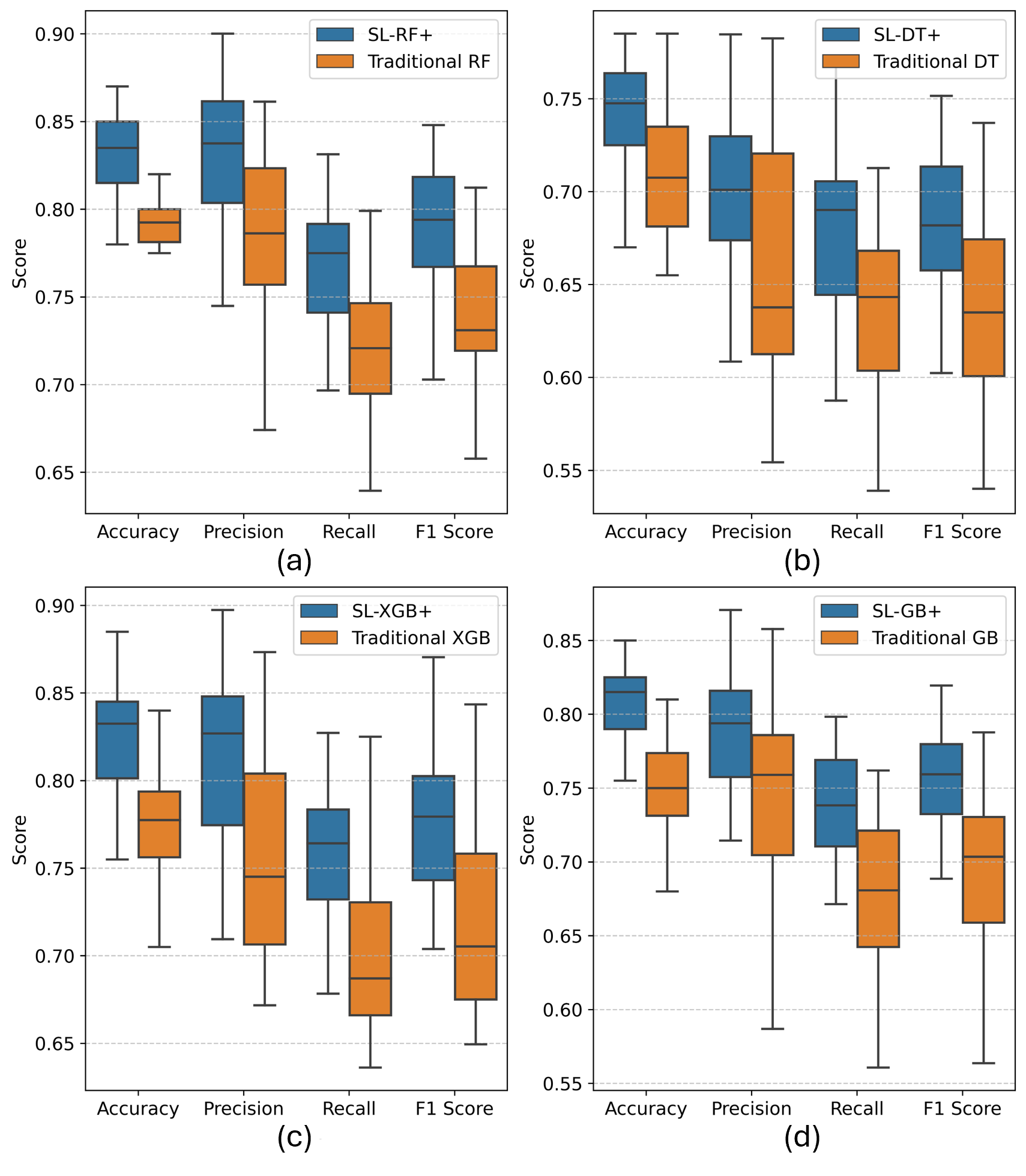}  
\caption{Boxplots of performance metrics across 30 runs for sequential models and their traditional ML counterparts: (a) RF, (b) DT, (c) XGB, and (d) GB}
\label{fig:boxplots_all_models}
\end{figure}

In data-scarce applications, collecting labeled data is often a costly and time-consuming process, making it impractical to train ML models on large datasets. This challenge is particularly relevant in AM domains like melt pool defect classification, where acquiring labeled experimental data is labor-intensive. The SL-RF+ framework has been designed to address this issue by focusing on data efficiency through selective sampling. This framework, when applied across various ML models, demonstrates superior performance over traditional ML methods that rely on fixed datasets. Figure~\ref{fig:boxplots_all_models} presents boxplots generated from 30 runs of four different ML models—Random Forest (RF), Decision Tree (DT), Gradient Boosting (GB), and Extreme Gradient Boosting (XGB)—in both our sequential (SL+) and traditional settings. These boxplots present a visual comparison of the performance achieved by this approach (SL+) versus traditional training approaches. From the plots, it is evident that SL+ variants consistently yield higher median scores and exhibit reduced variability compared to their traditional counterparts across all metrics. For instance, SL-RF+ shows an improvement in accuracy, precision, recall, and F1 score over traditional RF (Figure~\ref{fig:boxplots_all_models}a), highlighting the effectiveness of SL in enhancing model performance while minimizing data requirements. Similarly, the SL+ versions of XGB (Figure~\ref{fig:boxplots_all_models}b), GB (Figure~\ref{fig:boxplots_all_models}c), and DT (Figure~\ref{fig:boxplots_all_models}d) surpass their traditional counterparts, with noticeable improvements in precision and recall, underscoring the value of selective sampling in capturing the most informative data points. The average metrics for each model, as shown in Table \ref{tab:comparison_metrics}, further validate these observations. SL-RF+ achieves the highest average accuracy (0.8327), precision (0.8305), recall (0.7697), and F1 score (0.7896), setting a benchmark for model performance under data-scarce conditions. The SL versions of XGB, GB, and DT also exhibit significant improvements in each metric compared to their traditional versions, reinforcing the adaptability of this framework across different ML algorithms.  

\begin{table}[!htb]
\centering
\caption{Comparison of Average Metrics for Different Models}
\label{tab:comparison_metrics}
\small 
\begin{tabular}{@{}lcccc@{}}
\toprule
\textbf{Model} & \textbf{Accuracy} & \textbf{Precision} & \textbf{Recall} & \textbf{F1 Score} \\
\midrule
SL-RF+ & 0.8327 & 0.8305 & 0.7697 & 0.7896 \\
Traditional RF & 0.7877 & 0.7745 & 0.7168 & 0.7334 \\
SL-XGB+ & 0.8240 & 0.8156 & 0.7581 & 0.7757 \\
Traditional XGB & 0.7757 & 0.7511 & 0.6993 & 0.7141 \\
SL-GB+ & 0.8080 & 0.7927 & 0.7404 & 0.7573 \\
Traditional GB & 0.7507 & 0.7461 & 0.6757 & 0.6949 \\
SL-DT+ & 0.7512 & 0.7103 & 0.6911 & 0.6929 \\
Traditional DT & 0.7105 & 0.6570 & 0.6407 & 0.6394 \\
\bottomrule
\end{tabular}
\end{table}

\section{Conclusion}

In this study, a novel SL framework, SL-RF+, is presented for efficient and robust melt pool defect classification in L-PBF processes. This framework integrates an RF classifier with LCS and Sobol sequence-based synthetic sampling, benefiting from each component's unique strengths to achieve superior performance with a limited amount of labeled data. SL-RF+ offers a strategic approach to data efficiency, iteratively selecting the most informative samples to enhance classification accuracy and robustness in a data-scarce environment. The results demonstrate that SL-RF+ outperforms traditional RF and other ML models, achieving higher accuracy, precision, recall, and F1 scores with significantly fewer labeled samples. Our experiments show that the SL-RF+ framework reached optimal performance before 150 iterations, whereas the traditional RF model require a fully labeled set of 275 samples to achieve similar metrics. This efficiency highlights the value of incorporating uncertainty-based sampling strategies and quasi-random synthetic sample generation which enables this framework to effectively learn critical feature space regions. The class-wise performance analysis further validates SL-RF+’s ability to address class imbalances, particularly for rare defect types like balling and keyhole.

While SL-RF+ shows clear advantages over conventional methods, we acknowledge that the performance improvements are moderate due to the limitations of the pre-existing dataset. The experimental data used in this study has been collected beforehand, limiting the SL technique to a fixed set of data points. In a real-world application, where experiments should be conducted in response to the model's sampling suggestions, the performance gains of SL-RF+ would likely be more pronounced because the model would actively select and label samples that provide the highest potential improvement. This adaptivity  enables the SL-RF+ framework to achieve greater accuracy and reliability by accessing high-uncertainty samples as needed. Nevertheless, this study demonstrates that SL-RF+ can achieve strong performance even under data-scarce conditions by optimizing the use of available data. This approach sets a new benchmark for data efficiency in melt pool defect classification and offers a scalable solution for broader AM contexts, where selective sampling and SL frameworks can substantially reduce labeling costs while enhancing model robustness and reliability.

\section*{Acknowledgements}

The authors would like to thank the Department of Industrial and Management Systems Engineering (IMSE) for their invaluable support throughout this research.

 \bibliographystyle{elsarticle-num} 
 \bibliography{cas-refs}

\begin{thebibliography}{10}
\expandafter\ifx\csname url\endcsname\relax
  \def\url#1{\texttt{#1}}\fi
\expandafter\ifx\csname urlprefix\endcsname\relax\def\urlprefix{URL }\fi
\expandafter\ifx\csname href\endcsname\relax
  \def\href#1#2{#2} \def\path#1{#1}\fi

\bibitem{ngo2018additive}
T.~D. Ngo, A.~Kashani, G.~Imbalzano, K.~T. Nguyen, D.~Hui, Additive manufacturing (3d printing): A review of materials, methods, applications and challenges, Composites Part B: Engineering 143 (2018) 172--196.

\bibitem{ford2016additive}
S.~Ford, M.~Despeisse, Additive manufacturing and sustainability: an exploratory study of the advantages and challenges, Journal of cleaner Production 137 (2016) 1573--1587.

\bibitem{kumar2023metal}
G.~R. Kumar, M.~Sathishkumar, M.~Vignesh, M.~Manikandan, G.~Rajyalakshmi, R.~Ramanujam, N.~Arivazhagan, Metal additive manufacturing of commercial aerospace components--a comprehensive review, Proceedings of the Institution of Mechanical Engineers, Part E: Journal of Process Mechanical Engineering 237~(2) (2023) 441--454.

\bibitem{zhao2023direct}
N.~Zhao, M.~Parthasarathy, S.~Patil, D.~Coates, K.~Myers, H.~Zhu, W.~Li, Direct additive manufacturing of metal parts for automotive applications, Journal of Manufacturing Systems 68 (2023) 368--375.

\bibitem{zhang2023application}
Q.~Zhang, Y.~Guan, Application of metal additive manufacturing in oral dentistry, Current Opinion in Biomedical Engineering 25 (2023) 100441.

\bibitem{wang2020machine}
C.~Wang, X.~P. Tan, S.~B. Tor, C.~Lim, Machine learning in additive manufacturing: State-of-the-art and perspectives, Additive Manufacturing 36 (2020) 101538.

\bibitem{markl2016multiscale}
M.~Markl, C.~K{\"o}rner, Multiscale modeling of powder bed--based additive manufacturing, Annual Review of Materials Research 46~(1) (2016) 93--123.

\bibitem{scime2019using}
L.~Scime, J.~Beuth, Using machine learning to identify in-situ melt pool signatures indicative of flaw formation in a laser powder bed fusion additive manufacturing process, Additive Manufacturing 25 (2019) 151--165.

\bibitem{ye2023predictions}
J.~Ye, A.~Bab-Hadiashar, R.~Hoseinnezhad, N.~Alam, A.~Vargas-Uscategui, M.~Patel, I.~Cole, Predictions of in-situ melt pool geometric signatures via machine learning techniques for laser metal deposition, International Journal of Computer Integrated Manufacturing 36~(9) (2023) 1345--1361.

\bibitem{herzog2024process}
T.~Herzog, M.~Brandt, A.~Trinchi, A.~Sola, A.~Molotnikov, Process monitoring and machine learning for defect detection in laser-based metal additive manufacturing, Journal of Intelligent Manufacturing 35~(4) (2024) 1407--1437.

\bibitem{gordon2020defect}
J.~V. Gordon, S.~P. Narra, R.~W. Cunningham, H.~Liu, H.~Chen, R.~M. Suter, J.~L. Beuth, A.~D. Rollett, Defect structure process maps for laser powder bed fusion additive manufacturing, Additive Manufacturing 36 (2020) 101552.

\bibitem{wang2022data}
Z.~Wang, W.~Yang, Q.~Liu, Y.~Zhao, P.~Liu, D.~Wu, M.~Banu, L.~Chen, Data-driven modeling of process, structure and property in additive manufacturing: A review and future directions, Journal of Manufacturing Processes 77 (2022) 13--31.

\bibitem{kouraytem2021modeling}
N.~Kouraytem, X.~Li, W.~Tan, B.~Kappes, A.~D. Spear, Modeling process--structure--property relationships in metal additive manufacturing: a review on physics-driven versus data-driven approaches, Journal of Physics: Materials 4~(3) (2021) 032002.

\bibitem{akbari2022meltpoolnet}
P.~Akbari, F.~Ogoke, N.-Y. Kao, K.~Meidani, C.-Y. Yeh, W.~Lee, A.~B. Farimani, Meltpoolnet: Melt pool characteristic prediction in metal additive manufacturing using machine learning, Additive Manufacturing 55 (2022) 102817.

\bibitem{meng2020machine}
L.~Meng, B.~McWilliams, W.~Jarosinski, H.-Y. Park, Y.-G. Jung, J.~Lee, J.~Zhang, Machine learning in additive manufacturing: a review, Jom 72 (2020) 2363--2377.

\bibitem{johnson2020invited}
N.~Johnson, P.~Vulimiri, A.~To, X.~Zhang, C.~Brice, B.~Kappes, A.~Stebner, Invited review: Machine learning for materials developments in metals additive manufacturing, Additive Manufacturing 36 (2020) 101641.

\bibitem{du2020effects}
A.~Du~Plessis, I.~Yadroitsava, I.~Yadroitsev, Effects of defects on mechanical properties in metal additive manufacturing: A review focusing on x-ray tomography insights, Materials \& Design 187 (2020) 108385.

\bibitem{mukherjee2018mitigation}
T.~Mukherjee, T.~DebRoy, Mitigation of lack of fusion defects in powder bed fusion additive manufacturing, Journal of Manufacturing Processes 36 (2018) 442--449.

\bibitem{raihan2024augmented}
A.~S. Raihan, H.~Khosravi, T.~H. Bhuiyan, I.~Ahmed, An augmented surprise-guided sequential learning framework for predicting the melt pool geometry, Journal of Manufacturing Systems 75 (2024) 56--77.

\bibitem{zhang2021prediction}
Z.~Zhang, Z.~Liu, D.~Wu, Prediction of melt pool temperature in directed energy deposition using machine learning, Additive Manufacturing 37 (2021) 101692.

\bibitem{mojumder2023linking}
S.~Mojumder, Z.~Gan, Y.~Li, A.~Al~Amin, W.~K. Liu, Linking process parameters with lack-of-fusion porosity for laser powder bed fusion metal additive manufacturing, Additive Manufacturing 68 (2023) 103500.

\bibitem{rovzanec2024active}
J.~M. Ro{\v{z}}anec, L.~Bizjak, E.~Trajkova, P.~Zajec, J.~Keizer, B.~Fortuna, D.~Mladeni{\'c}, Active learning and novel model calibration measurements for automated visual inspection in manufacturing, Journal of Intelligent Manufacturing 35~(5) (2024) 1963--1984.

\bibitem{chen2024meltpoolgan}
H.~Chen, X.~Liu, X.~Liu, P.~Witherell, Meltpoolgan: Melt pool prediction from path-level thermal history, Additive Manufacturing 84 (2024) 104095.

\bibitem{snow2021toward}
Z.~Snow, B.~Diehl, E.~W. Reutzel, A.~Nassar, Toward in-situ flaw detection in laser powder bed fusion additive manufacturing through layerwise imagery and machine learning, Journal of Manufacturing Systems 59 (2021) 12--26.

\bibitem{zoller2023numerical}
C.~Z{\"o}ller, N.~Adams, S.~Adami, Numerical investigation of balling defects in laser-based powder bed fusion of metals with inconel 718, Additive Manufacturing 73 (2023) 103658.

\bibitem{yang2022quality}
G.~Yang, Y.~Xie, S.~Zhao, L.~Qin, X.~Wang, B.~Wu, Quality control: Internal defects formation mechanism of selective laser melting based on laser-powder-melt pool interaction: A review, Chinese Journal of Mechanical Engineering: Additive Manufacturing Frontiers 1~(3) (2022) 100037.

\bibitem{clijsters2014situ}
S.~Clijsters, T.~Craeghs, S.~Buls, K.~Kempen, J.-P. Kruth, In situ quality control of the selective laser melting process using a high-speed, real-time melt pool monitoring system, The International Journal of Advanced Manufacturing Technology 75 (2014) 1089--1101.

\bibitem{liu2020study}
W.~Liu, C.~Chen, S.~Shuai, R.~Zhao, L.~Liu, X.~Wang, T.~Hu, W.~Xuan, C.~Li, J.~Yu, et~al., Study of pore defect and mechanical properties in selective laser melted ti6al4v alloy based on x-ray computed tomography, Materials Science and Engineering: A 797 (2020) 139981.

\bibitem{angelastro2021integrated}
A.~Angelastro, S.~L. Campanelli, An integrated analytical model for the forecasting of the molten pool dimensions in selective laser melting, Laser Physics 32~(2) (2021) 026001.

\bibitem{soundararajan2021review}
B.~Soundararajan, D.~Sofia, D.~Barletta, M.~Poletto, Review on modeling techniques for powder bed fusion processes based on physical principles, Additive Manufacturing 47 (2021) 102336.

\bibitem{guo2022machine}
S.~Guo, M.~Agarwal, C.~Cooper, Q.~Tian, R.~X. Gao, W.~G. Grace, Y.~Guo, Machine learning for metal additive manufacturing: Towards a physics-informed data-driven paradigm, Journal of Manufacturing Systems 62 (2022) 145--163.

\bibitem{ho2021dlam}
S.~Ho, W.~Zhang, W.~Young, M.~Buchholz, S.~Al~Jufout, K.~Dajani, L.~Bian, M.~Mozumdar, Dlam: Deep learning based real-time porosity prediction for additive manufacturing using thermal images of the melt pool, IEEE Access 9 (2021) 115100--115114.

\bibitem{wang2024traditional}
H.~Wang, B.~Li, S.~Zhang, F.~Xuan, Traditional machine learning and deep learning for predicting melt-pool cross-sectional morphology of laser powder bed fusion additive manufacturing with thermographic monitoring, Journal of Intelligent Manufacturing (2024) 1--26.

\bibitem{abranovic2024melt}
B.~Abranovic, S.~Sarkar, E.~Chang-Davidson, J.~Beuth, Melt pool level flaw detection in laser hot wire directed energy deposition using a convolutional long short-term memory autoencoder, Additive Manufacturing 79 (2024) 103843.

\bibitem{choudhary2022recent}
K.~Choudhary, B.~DeCost, C.~Chen, A.~Jain, F.~Tavazza, R.~Cohn, C.~W. Park, A.~Choudhary, A.~Agrawal, S.~J. Billinge, et~al., Recent advances and applications of deep learning methods in materials science, npj Computational Materials 8~(1) (2022) 59.

\bibitem{delgado2021deep}
J.~M.~D. Delgado, L.~Oyedele, Deep learning with small datasets: using autoencoders to address limited datasets in construction management, Applied Soft Computing 112 (2021) 107836.

\bibitem{kwon2020deep}
O.~Kwon, H.~G. Kim, M.~J. Ham, W.~Kim, G.-H. Kim, J.-H. Cho, N.~I. Kim, K.~Kim, A deep neural network for classification of melt-pool images in metal additive manufacturing, Journal of Intelligent Manufacturing 31 (2020) 375--386.

\bibitem{dasari2020melt}
S.~K. Dasari, A.~Cheddad, J.~Palmquist, Melt-pool defects classification for additive manufactured components in aerospace use-case, in: 2020 7th International Conference on Soft Computing \& Machine Intelligence (ISCMI), IEEE, 2020, pp. 249--254.

\bibitem{gaikwad2022multi}
A.~Gaikwad, R.~J. Williams, H.~de~Winton, B.~D. Bevans, Z.~Smoqi, P.~Rao, P.~A. Hooper, Multi phenomena melt pool sensor data fusion for enhanced process monitoring of laser powder bed fusion additive manufacturing, Materials \& Design 221 (2022) 110919.

\bibitem{mao2023continuous}
Z.~Mao, W.~Feng, H.~Ma, Y.~Yang, J.~Zhou, S.~Liu, Y.~Liu, P.~Hu, K.~Zhao, H.~Xie, et~al., Continuous online flaws detection with photodiode signal and melt pool temperature based on deep learning in laser powder bed fusion, Optics \& Laser Technology 158 (2023) 108877.

\bibitem{mahato2022detecting}
V.~Mahato, M.~A. Obeidi, D.~Brabazon, P.~Cunningham, Detecting voids in 3d printing using melt pool time series data, Journal of Intelligent Manufacturing (2022) 1--8.

\bibitem{gui2022detection}
Y.~Gui, K.~Aoyagi, H.~Bian, A.~Chiba, Detection, classification and prediction of internal defects from surface morphology data of metal parts fabricated by powder bed fusion type additive manufacturing using an electron beam, Additive Manufacturing 54 (2022) 102736.

\bibitem{wang2023gaussian}
Q.~Wang, X.~Lin, X.~Duan, R.~Yan, J.~Y.~H. Fuh, K.~Zhu, Gaussian process classification of melt pool motion for laser powder bed fusion process monitoring, Mechanical Systems and Signal Processing 198 (2023) 110440.

\bibitem{yang2023defect}
W.~Yang, Y.~Qiu, W.~Liu, X.~Qiu, Q.~Bai, Defect prediction in laser powder bed fusion with the combination of simulated melt pool images and thermal images, Journal of Manufacturing Processes 106 (2023) 214--222.

\bibitem{smoqi2022monitoring}
Z.~Smoqi, A.~Gaikwad, B.~Bevans, M.~H. Kobir, J.~Craig, A.~Abul-Haj, A.~Peralta, P.~Rao, Monitoring and prediction of porosity in laser powder bed fusion using physics-informed meltpool signatures and machine learning, Journal of Materials Processing Technology 304 (2022) 117550.

\bibitem{sato2024identification}
M.~M. Sato, V.~W. Wong, H.~Yeung, P.~Witherell, K.~H. Law, Identification and interpretation of melt pool shapes in laser powder bed fusion with machine learning, Smart and Sustainable Manufacturing Systems 8~(1) (2024) 1--23.

\bibitem{gu2024deep}
Z.~Gu, K.~Mani~Krishna, M.~Parsazadeh, S.~Sharma, A.~Manjunath, H.~Tran, S.~Fu, N.~B. Dahotre, Deep learning-based melt pool and porosity detection in components fabricated by laser powder bed fusion, Progress in Additive Manufacturing (2024) 1--18.

\bibitem{khanzadeh2019situ}
M.~Khanzadeh, S.~Chowdhury, M.~A. Tschopp, H.~R. Doude, M.~Marufuzzaman, L.~Bian, In-situ monitoring of melt pool images for porosity prediction in directed energy deposition processes, IISE Transactions 51~(5) (2019) 437--455.

\bibitem{ertay2021process}
D.~S. Ertay, M.~A. Naiel, M.~Vlasea, P.~Fieguth, Process performance evaluation and classification via in-situ melt pool monitoring in directed energy deposition, CIRP Journal of Manufacturing Science and Technology 35 (2021) 298--314.

\bibitem{zhang2018extraction}
Y.~Zhang, G.~S. Hong, D.~Ye, K.~Zhu, J.~Y. Fuh, Extraction and evaluation of melt pool, plume and spatter information for powder-bed fusion am process monitoring, Materials \& Design 156 (2018) 458--469.

\bibitem{wu2024situ}
Q.~Wu, F.~Yang, C.~Lv, C.~Liu, W.~Tang, J.~Yang, In-situ quality intelligent classification of additively manufactured parts using a multi-sensor fusion based melt pool monitoring system, Additive Manufacturing Frontiers 3~(3) (2024) 200153.

\bibitem{petrik2023meltpoolgan}
J.~Petrik, B.~Kavas, M.~Bambach, Meltpoolgan: Auxiliary classifier generative adversarial network for melt pool classification and generation of laser power, scan speed and scan direction in laser powder bed fusion, Additive Manufacturing 78 (2023) 103868.

\bibitem{ahmed2024toward}
I.~Ahmed, S.~T. Bukkapatnam, B.~Botcha, Y.~Ding, Toward futuristic autonomous experimentation—a surprise-reacting sequential experiment policy, IEEE Transactions on Automation Science and Engineering (2024).

\bibitem{van2021active}
G.~J. van Houtum, M.~L. Vlasea, Active learning via adaptive weighted uncertainty sampling applied to additive manufacturing, Additive Manufacturing 48 (2021) 102411.

\bibitem{dasari2021active}
S.~K. Dasari, A.~Cheddad, L.~Lundberg, J.~Palmquist, Active learning to support in-situ process monitoring in additive manufacturing, in: 2021 20th IEEE international conference on machine learning and applications (ICMLA), IEEE, 2021, pp. 1168--1173.

\bibitem{chepiga2023process}
T.~Chepiga, P.~Zhilyaev, A.~Ryabov, A.~P. Simonov, O.~N. Dubinin, D.~G. Firsov, Y.~O. Kuzminova, S.~A. Evlashin, Process parameter selection for production of stainless steel 316l using efficient multi-objective bayesian optimization algorithm, Materials 16~(3) (2023) 1050.

\bibitem{liu2022nonparametric}
J.~Liu, J.~Ye, F.~Momin, X.~Zhang, A.~Li, Nonparametric bayesian framework for material and process optimization with nanocomposite fused filament fabrication, Additive Manufacturing 54 (2022) 102765.

\bibitem{parmar2019review}
A.~Parmar, R.~Katariya, V.~Patel, A review on random forest: An ensemble classifier, in: International conference on intelligent data communication technologies and internet of things (ICICI) 2018, Springer, 2019, pp. 758--763.

\bibitem{agrawal2021active}
A.~Agrawal, S.~Tripathi, M.~Vardhan, Active learning approach using a modified least confidence sampling strategy for named entity recognition, Progress in Artificial Intelligence 10~(2) (2021) 113--128.

\bibitem{renardy2021sobol}
M.~Renardy, L.~R. Joslyn, J.~A. Millar, D.~E. Kirschner, To sobol or not to sobol? the effects of sampling schemes in systems biology applications, Mathematical biosciences 337 (2021) 108593.

\end{thebibliography}





\end{document}